# Fake News Detection: a comparison between available Deep Learning techniques in vector space

Lovedeep Singh
Computer Science & Engineering
Punjab Engineering College
Chandigarh, India
masterlovedeep.singh97@gmail.com

*Abstract*—Fake News Detection is an essential problem in the field of Natural Language Processing. The benefits of an effective solution in this area are manifold for the goodwill of society. On a surface level, it broadly matches with the general problem of text classification. Researchers have proposed various approaches to tackle fake news using simple as well as some complex techniques. In this paper, we try to make a comparison between the present Deep Learning techniques by representing the news instances in some vector space using a combination of common mathematical operations with available vector space representations. We do a number of experiments using various combinations and permutations. Finally, we conclude with a sound analysis of the results and evaluate the reasons for such results.

## I. INTRODUCTION

Fake News Detection is a classic problem of classifying a piece of text as true or false. We can tackle fake news by using some mathematical treatment on the text and leaving it on the Machine Learning or Deep Learning models to do the rest. Another way is to cleverly figure out the distinguishing factor(s) between true and fake instances and use those factors to train the models. The former approach is more straightforward and can give different results for a different combination of models and vector representations. The latter approach is dependent on the domain and is limited by a restricted amount of data of a particular segment playing with those specific factors. Although, if can to extract those factors, use them for training the machine learning models, and test them on a dataset playing with those features, the accuracy is bound to be high. But unfortunately, the latter approach is not generic, requires domain-specific knowledge, some specific sort of pre-processing, and is unlikely to scale efficiently. In this paper, we focus on the former approach and try to argue on different Deep Learning techniques using various vector space representations. We focus on the three models of Deep Learning, Artificial Neural Networks, Convolutional Neural Networks, and Recurrent Neural Networks.

## II. RELATED WORK

Researchers[1][2] have tried fake news detection using linguistic features other than vector space representations on fundamental machine learning models like SVM. With the advent of neural networks, an increase in computes power and a lot of data at disposal, researchers[3][4][5] have even tried various combinations using Deep Learning techniques to overcome fake news. Researchers[6] have also tried ensemble techniques either by voting algorithms or by stacking techniques on top of each other. Most of the work has been done to address the shortcoming of previous techniques and improve upon them. We feel that certain techniques can be good in one circumstance and bad in other, there are can be a lot of factors that help a particular technique in certain scenarios and hinder it in other ones. With this paper, we work on experiments to address and analyse such circumstances.

## III. DATASET

We use two different datasets in our experiments. One of the datasets is the LIAR[7] dataset. It is a document database; in a sense, it contains 12 columns in each tuple instead of simple text and labels, and it divides the tuples into six categories instead of simple truth and false. This dataset contains simple statements of one line and details such as the speaker, event, speaker history of truthfulness, and some other columns. The other dataset we use is the public Fake News dataset available on Kaggle[1]; this is more like a traditional dataset, unlike the former document dataset. It contains mainly the title, text, author, and label in each tuple. The main difference between the datasets other than the number of attributes in each tuple is the length of each record. LIAR dataset contains short statements, whereas the latter dataset contains the complete long-form text of about 20 lines in each tuple. This varying nature between the two datasets will help us argue the efficacy of the vector space representations between two extremities along with their combinations with different Deep Learning models. Both of these datasets are well balanced and contain an almost equal number of records in both true and false categories. The LIAR dataset contains around 12k records, whereas the Kaggle dataset contains about 20k records.

### A. Initial Data Preprocessing

In the LIAR dataset, we merge the three categories bending towards truth as true and three false categories bending towards fallacy as false to increase the number of tuples in each category. We do not use all the 12 columns; we use only the statement and the label. Our second dataset is the Fake News dataset available on Kaggle. It is a robust and big dataset. LIAR dataset contains simple statements, whereas the latter dataset contains title and text, and each text is a long paragraph of about twenty lines. In the latter dataset, we combine title and text as one column of the tuple. Finally, in both the datasets, we use only two columns in each tuple, content, and label.

---

[1] https://www.kaggle.com/c/fake-news/data

## B. Basic Data Preprocessing before vectorization

In all of our experiments, we pre-process our data before passing it for training to the Deep Learning models. We remove the stop-words and use the Porter Stemming algorithm to stem the words back to the root word. Other than this, some specific tweaks have been done depending on the vector space representation and the model, which have been discussed under the subsequent sections.

## IV. VECTOR SPACE REPRESENTATION

We use total 4 vector space representations in our experiments. These are one-hot encoding, TFIDF, Word2Vec, and Doc2Vec. In our models, the smallest representative unit generated in the vector space is for a word, researchers have even proposed character level encoding for generating space vectors. Researchers[8] elaborate that this is successful on much larger datasets, but TFIDF and other basic techniques perform better on smaller datasets. So, we stick to the word being the smallest representative unit in vector space. To avoid any bias, wherever possible, we split the dataset before generating vector space representations and do separate transformations on both train and test datasets.

### A. Word2Vec

We train genism[2] model to generate vector representations for words in our corpus instead of using pre-trained models. Leaving Word2Vec, other representations directly give space vector for the complete paragraph. In Word2Vec, we need to combine the space vectors of the words in the sentence using some technique to generate the entire paragraph's representation. One way to combine word vectors is by stacking the vector of the next word below the vector of the previous word; this will give a two-dimensional matrix for each paragraph. Although this looks promising, to be fair with other representations along one axis, we do not use such a method. Instead, we take the average of all the word vectors to get the complete paragraph vector representation. Our word vectors are of dimension 100; we have kept the minimum count to be 1, i.e., we have generated representation even for one time occurring words in the dataset.

### B. One-hot encoding

This is perhaps the most straightforward representation in all the vector space models. We keep the vocabulary length sufficiently 5000 to cover all the words. We keep the sentence length to 20 and do padding at the beginning wherever required.

### C. Doc2Vec

Doc2Vec is an extension to Word2Vec capable of representing the complete paragraph and preserving the ordering of the words. We use the genism[3] model to train our own Doc2Vec and keep our vectors of dimension 300.

### D. TFIDF

This is the simplest vector space model after one-hot encodings. Its potential to reflect the importance of the word proportional to its frequency and not letting the too common words dominate has been proven effective in bringing out the essential words in the document. This was the most intense representation for our Deep Learning models due to high dimensionality. In the LIAR dataset, for each statement, we had a vector of dimension 11915. In the Fake News dataset of Kaggle, for each paragraph, we got a vector of dimension 157102.

## V. DEEP LEARNING MODELS

We use the three main models of Deep Learning. The first model is an Artificial Neural Network, ANN form the basis of Deep Learning. The second model is the Convolutional Neural Network. Initially developed for images and areas related to Computer Vision, recently, CNNs have been used in NLP problems as well. Our last model is Recurrent Neural Network. RNNs are being used widely in NLP recently with promising results in areas such as text generation. Their capability to take into consideration previous output helps them to recognize patterns that are not captured well by simple ANNs.

### A. ANN

There is no standard rule for the number of hidden layers in an ANN. We have not followed any particular fixed pattern in our models, but we have mostly reduced the size of the output layer to half of the input layer until we culminate. The intuition was to keep the number of layers proportional to the size of the input. The higher the input vector's dimension, the more the number of layers we have in our neural network. The activation function in each layer is relu, and the activation function in the final layer is sigmoid.

### B. CNN

We use two 1D convolutional layers with a dropout of 0.5 after each layer and a final pooling layer before a dense layer for output. Again, there are no standard rules for the number of convolutional layers, the fraction of dropout, and the length of the convolutional window. We have used a window of length 5 with a stride of 1 unit. We have used a total of 128 filters in the first convolutional layer and 64 in the second. The activation function is relu in each convolutional layer and softmax in the final dense layer.

### C. RNN

There are various flavors to RNN, each trying a different approach to tackle the diminishing gradient problem. LSTMs are an intuitive solution to a diminishing gradient. We have used Bi-LSTMs instead of LSTMs. Research[6] has shown that Bi-LSTMs perform better. We have one layer with 100 neurons and a dropout of 0.3 before a final dense layer with a sigmoid activation function.

## VI. METHODOLOGY

The figure below depicts the complete methodology used to conduct the experiments using a flow-diagram. Starting from cleaning the data, initial pre-processing to vectorization to final deep learning models on top of these vector representations. Most of the steps are identical in both the LIAR and Kaggle dataset except some differences in the initial steps of the data pre-processing. These have been neatly depicted in the Fig 1. and Fig 2. for LIAR and Kaggle datasets respectively. Finally we get the accuracies of various combinations of vector techniques and deep learning models and both the datasets and analyse the results.

---

[2] https://radimrehurek.com/gensim/models/word2vec.html

[3] https://radimrehurek.com/gensim/models/doc2vec.html

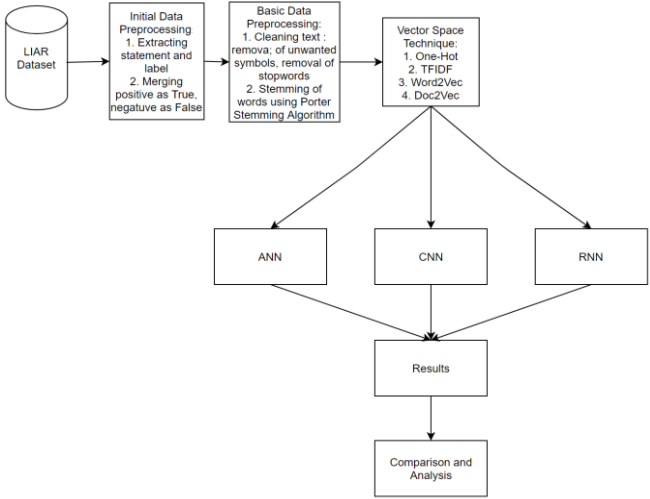

Fig. 1. Flow Diagram for LIAR datset

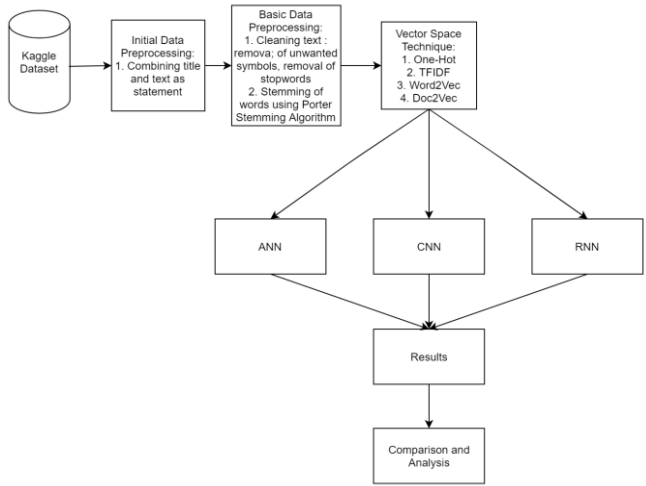

Fig. 2. Flow Diagram for Kaggle dataset

## VII. RESULTS

Tables below show the results of the experiments conducted using various Deep Learning Models on Vector Space Representations. Each Table shows the Accuracy (%) of the Model used on the particular dataset using a specific Vector Space technique.

TABLE I. ANN

| Dataset | Vector Space Representation | | | |
|---|---|---|---|---|
| | *One-Hot* | *TFIDF* | *Word2Vect* | *Doc2Vec* |
| LIAR | 54.17% | 54.17% | 54.17% | 54.17% |
| Kaggle | 52.03% | 96.59% | 70.06% | 79.66% |

TABLE II. CNN

| Dataset | Vector Space Representation | | | |
|---|---|---|---|---|
| | *One-Hot* | *TFIDF* | *Word2Vect* | *Doc2Vec* |
| LIAR | 54.17% | 59.82% | 55.63% | 54.17% |
| Kaggle | 53.96% | 96.89% | 79.29% | 80.89% |

TABLE III. RNN

| Dataset | Vector Space Representation | | | |
|---|---|---|---|---|
| | *One-Hot* | *TFIDF* | *Word2Vect* | *Doc2Vec* |
| LIAR | 58.13% | 59.82% | 45.83% | 54.17% |
| Kaggle | 84.62% | 96.81% | 69.55% | 82.95% |

## VIII. CONCLUSION

Although accuracy is not always the best evaluation metric, since datasets were balanced, a very high accuracy indeed means that the model performed well. CNN performed the best with TFIDF on the Kaggle dataset. All three models performed very well with TFIDF on the Kaggle dataset. We can infer that TFIDF is an effective approach when the dataset has large paragraphs for each news instance. The LIAR dataset is a document dataset in original sense and contains very short statements. It did not perform well in any of the models and vector space combinations. The nature of the dataset can explain this; the statements were short, were little related in vector space, i.e., they did not have much similar words in them, and therefore did not generated very meaningful vector space representations. Coming to the Kaggle dataset and One-Hot encoding, it is surprising that only Bi-LSTM performed well among all three models. The nature of Bi-LSTM can explain this; it tries to learn the ordering of the words in both directions, therefore for Bi-LSTM, encoded words were easy to understand in context to their ordering in long paragraphs of Kaggle dataset.

For the same reason, in LIAR with One-Hot encoding, Bi-LSTM performed better than other models. It did not perform as well as Kaggle since the statements were short and little related. Word2Vec on LIAR did not perform well in any of the models. It is clear that by averaging the vectors for different words in LIAR's short statements, we lost the entire meaning. The surprising thing is that Bi-LSTM performed worst with Word2Vec on LIAR. Bi-LSTM tried to learn in a way similar to One-Hot encoding, but by averaging the vectors, we already had lost the essence and, therefore, learned the wrong relations. Coming to Doc2Vec, in the LIAR dataset, since the statements were short and little related, we did not get a meaningful Doc2Vec.

On the other hand, in the Kaggle dataset, since the paragraphs were long, we got a meaningful Doc2Vec representation and got decent accuracies in all the three models. Bi-LSTM performed better than other models with Doc2Vec on Kaggle dataset because of its bidirectional nature of learning. An-other surprise is that even though we lost the essence of the statements by averaging word vectors in the LIAR dataset, we still got some decent results in the case of the Kaggle dataset. One interesting thing is the better performance of CNN with Word2Vec on the Kaggle dataset relative to the other two models. The nature of Convolutional Neural Networks can explain this. Although we lost some essence of the paragraphs because of averaging of word vectors, the convolutional and pooling operations someway helped to bring some more meaning to these representations compared to the other two models similar to how CNN performs in image recognition.

Complete code and dataset are available at GitHub[4] for further experiments and analysis.

REFERENCES

[1] Pérez-Rosas, Verónica & Kleinberg, Bennett & Lefevre, Alexandra & Mihalcea, Rada. (2017). Automatic Detec-tion of Fake News.

[2] Singh, Vivek & Dasgupta, Rupanjal & Sonagra, Darshan & Raman, Karthik & Ghosh, Isha. (2017). Automated Fake News Detection Using Linguistic Analy-sis and Machine Learning. 10.13140/RG.2.2.16825.67687.

[3] Huang, Yin-Fu & Chen, Po-Hong. (2020). Fake News Detection Using an Ensemble Learning Model Based on Self-adaptive Harmony Search Algorithms. Expert Systems with Applications. 159. 113584. 10.1016/j.eswa.2020.113584.

[4] Bahad, Pritika & Saxena, Preeti & Kamal, Raj. (2020). Fake News Detection using Bi-directional LSTM-Recurrent Neural Network. Procedia Computer Science. 165. 74-82. 10.1016/j.procs.2020.01.072.

[5] Zhang, Jiawei, Limeng Cui and Yanjie Fu. "FakeDetector: Effective Fake News Detection with Deep Diffusive Neural Network." 2020 IEEE 36th International Conference on Data Engineering (ICDE) (2020): 1826-1829.

[6] Thorne, James & Chen, Mingjie & Myrianthous, Giorgos & Pu, Jiashu & Wang, Xiaoxuan & Vlachos, Andreas. (2017). Fake news stance detection using stacked ensemble of classifiers. 80-83. 10.18653/v1/W17-4214.

[7] Wang, William. (2017). "Liar, Liar Pants on Fire": A New Benchmark Dataset for Fake News Detection. 422-426. 10.18653/v1/P17-2067.

[8] Zhang, Xiang & Zhao, Junbo & Lecun, Yann. (2015). Character-level Convolutional Networks for Text Classification.

---

[4] https://github.com/singh-l/FNDLVS